\documentclass[lettersize,journal]{IEEEtran}

\usepackage[switch]{lineno}

\usepackage{amsmath,amsfonts,amsthm}

\usepackage{algorithmic}
\usepackage{algorithm}
\usepackage{array}
\usepackage[caption=false,font=normalsize,labelfont=sf,textfont=sf]{subfig}
\usepackage{textcomp}
\usepackage{stfloats}
\usepackage{url}
\usepackage{verbatim}
\usepackage{graphicx}
\usepackage{cite}
\usepackage{arydshln}
\usepackage{multirow}
\usepackage[normalem]{ulem}
\usepackage{makecell}
\usepackage{diagbox}

\usepackage{xcolor}
\usepackage{enumitem}
\usepackage{xspace}

\theoremstyle{definition}

\newtheorem*{problem*}{\textbf{Problem Definition}}

\newcommand{\model}{\emph{CrossHGL}\xspace}

\usepackage{pdfpages}
\usepackage{fancyhdr}

\begin{document}
\title{CrossHGL: A Text-Free Foundation Model for Cross-Domain Heterogeneous Graph Learning}

\author{
Xuanze Chen, 
Jiajun Zhou, 
Yadong Li,
Shanqing Yu, 
Qi Xuan, \IEEEmembership{Senior Member, IEEE}
\thanks{This work was supported in part by National Natural Science Foundation of China (No. 62503423), in part by the National Key Research and Development Program of China (No. 2025YFA1510900), in part by the Key Research and Development Program of Zhejiang Province (No. 2026C02A1233), in part by the Yangtze River Delta Science and Technology Innovation Community Joint Research Project (No. 2026ZY03003, No. 2025CSJGG01000). \emph{(Corresponding authors: Jiajun Zhou.)}}

\thanks{Xuanze Chen and Yadong Li are with the Institute of Cyberspace Security, Zhejiang University of Technology, Hangzhou 310023, China, and also with the Binjiang Cyberspace Security Institute of ZJUT, Hangzhou, 310056, China (e-mail: chenxuanze@zjut.edu.cn).}
\thanks{Jiajun Zhou, Shanqing Yu and Qi Xuan are with the Institute of Cyberspace Security, Zhejiang University of Technology, Hangzhou 310023, China, with the Binjiang Cyberspace Security Institute of ZJUT, Hangzhou, 310056, China, and with the Soovar Technologies Co., Ltd., Hangzhou 310056, China (e-mail: jjzhou@zjut.edu.cn).}
}

\markboth{Journal of \LaTeX\ Class Files,~Vol.~14, No.~8, August~2021}%
{Shell \MakeLowercase{\textit{et al.}}: A Sample Article Using IEEEtran.cls for IEEE Journals}


\maketitle



\begin{abstract}
Heterogeneous graph representation learning has attracted increasing attention, as heterogeneous graphs can model complex systems with diverse node and edge types. Despite recent progress, most existing methods have been developed in closed-world settings, in which the training and testing data share similar graph schemas and feature spaces, thereby limiting their cross-domain generalization. Recent graph foundation models have improved transferability; however, existing approaches either focus on homogeneous graphs, rely on domain-specific heterogeneous schemas, or depend on rich textual attributes for semantic alignment. Consequently, cross-domain heterogeneous graph learning in text-free and few-shot settings remains insufficiently explored. To address this problem, we propose \model, a foundation framework that explicitly preserves and transfers multi-relational structural semantics without relying on external textual supervision. Specifically, a semantic-preserving graph transformation strategy is designed to homogenize heterogeneous graphs while encoding heterogeneous interaction semantics into edge features. Based on the transformed graph, a prompt-aware multi-domain pre-training framework with a Tri-Prompt mechanism captures transferable knowledge from feature, edge, and structure perspectives through self-supervised graph contrastive learning. For target-domain adaptation, we further develop a parameter-efficient fine-tuning strategy that freezes the pre-trained backbone and performs few-shot classification via prompt composition and prototypical learning. Experiments on downstream node-level and graph-level tasks show that \model consistently outperforms representative pre-training \& fine-tuning baselines, yielding average relative improvements of 25.1\% and 7.6\% in Micro-F1 for node classification and graph classification, respectively, while remaining competitive under more challenging feature-degenerated settings.
\end{abstract}
    

\section{Introduction}
\begin{figure}[!htb]
    \centering
    \includegraphics[width=\linewidth]{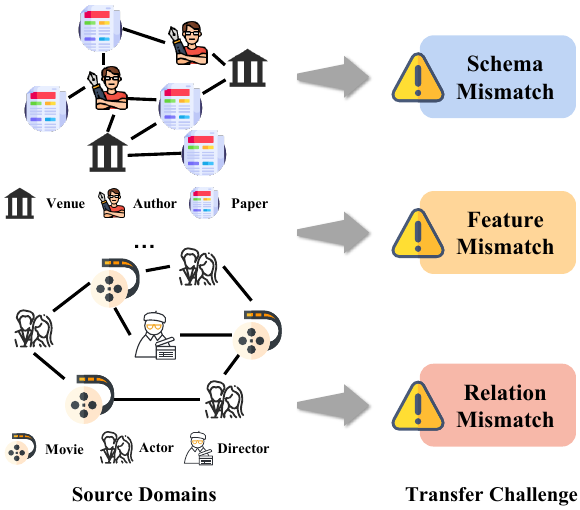}
    \caption{Challenges in cross-domain heterogeneous pre-training.}
    \label{fig:intro_motivation}
\end{figure}
Heterogeneous graphs, characterized by diverse types of nodes and edges, have become an important data structure for modeling complex interactive systems in the real world~\cite{jin2024enhancing,hu2018leveraging,jin2022heterogeneous,liu2018heterogeneous}. Over the past few years, Heterogeneous Graph Neural Networks (HGNNs) have achieved strong performance in learning expressive node and graph representations~\cite{RGCN,HGT}. Despite their effectiveness, conventional HGNNs inherently operate under a closed-world assumption, meaning they require the training and testing data to share strictly identical graph schemas and feature spaces, which substantially limits their generalization to unseen domains~\cite{zhang2022few,fewshotonG,dai2022graph}.

To alleviate the heavy reliance on massive labeled data, the pre-training and fine-tuning paradigm has recently stimulated the development of graph foundation models~\cite{GPPT,GraphPrompt,jiao2025hgmp}. Existing studies have advanced along several main directions, including foundation models for homogeneous graphs~\cite{liu2024one,sun2023all}, foundation models for heterogeneous graphs within specific domains~\cite{hepa,hgprompt}, and Large Language Model (LLM) enhanced heterogeneous graph foundation models~\cite{higpt,LMCH,jia2025hetgcot}. However, extending foundation models to cross-domain heterogeneous graphs without external text assistance remains challenging. A major difficulty lies in the severe heterogeneity shift across domains, where graph schemas, interaction patterns, and feature dimensions may differ substantially~\cite{yu2026evaluating}. Many existing heterogeneous foundation models are still tightly coupled with specific meta-paths or predefined relation types~\cite{xu2025heterogeneous}, which limits their applicability in more general cross-domain settings. As illustrated in Fig.~\ref{fig:intro_motivation}, when a model is pre-trained across multiple source heterogeneous graphs, schema mismatch, feature mismatch, and relation mismatch naturally arise across domains, making direct cross-domain transfer considerably more difficult.

Recently, several studies have attempted to unify cross-domain heterogeneous graphs by integrating LLMs~\cite{higpt,LMCH}. By aligning raw data into a shared textual semantic space, these LLM-enhanced models can alleviate structural mismatches across domains. Nevertheless, their effectiveness fundamentally depends on the availability of rich textual attributes. In many safety-critical and privacy-sensitive domains, such as financial transaction monitoring~\cite{huang2022dgraph}, network anomaly detection~\cite{liu2021anomaly} and molecular property prediction~\cite{ni2025robust} graph features are purely numerical, categorical, or anonymized, resulting in a text-free setting. In such scenarios, LLM-based semantic alignment is no longer applicable. Moreover, although text-free cross-domain foundation models have been explored for homogeneous graphs~\cite{GCOPE,SAMGPT}, directly applying them to heterogeneous graphs usually only aligns feature spaces and simplified neighborhood structures, while failing to adequately preserve the rich multi-relational semantics embedded in heterogeneous interactions. Consequently, robust knowledge transfer for text-free heterogeneous graphs under extreme few-shot constraints remains under-explored.

To address these limitations, a novel and adaptable framework, denoted as \model, is proposed for text-free, few-shot, cross-domain heterogeneous graph learning. Rather than relying on external textual semantics or indiscriminately flattening network topologies, the proposed framework explicitly decouples and transfers the inherent structural and topological semantics within graphs. The overall architecture consists of three phases. First, a semantic-preserving graph transformation strategy is designed to reduce heterogeneity shifts while retaining informative interaction patterns. This phase leverages SVD-based feature alignment and automated meta-pattern mining to homogenize heterogeneous topologies while compressing multi-relational context into semantics-enriched edges. Second, a prompt-aware multi-domain pre-training mechanism is introduced. By integrating a Tri-Prompt system, including feature, edge, and structure prompts, with a shared Graph Neural Network (GNN) backbone, the framework captures transferable representations through self-supervised graph contrastive learning. Finally, a parameter-efficient fine-tuning strategy is deployed for the target domain. By freezing the pre-trained backbone, the framework employs an attention-based prompt composition to adapt prior knowledge to novel domains, enabling few-shot classification via non-parametric prototypical networks.
The primary contributions of this work are consolidated as follows:
\begin{itemize}[leftmargin=10pt,nosep]
    \item It proposes \model, a foundation framework for text-free, few-shot, and cross-domain learning on heterogeneous graphs, avoiding the reliance on LLMs and reducing the semantic loss caused by direct topology flattening.
    \item A semantic-preserving graph transformation strategy is introduced, which couples SVD-based feature alignment with automated meta-pattern mining to unify disparate graph schemas into semantics-enriched homogeneous topologies.
    \item A parameter-efficient Tri-Prompt architecture is devised to decouple multi-dimensional graph semantics during pre-training, alongside an attention-based prompt composition mechanism that enables few-shot adaptation on a frozen GNN backbone. Extensive experiments on diverse real-world heterogeneous graph datasets demonstrate the effectiveness and transferability of the proposed framework.
\end{itemize}

\section{Related Work}
\subsection{Foundation Models for Homogeneous Graphs}
Recent years have witnessed the rapid development of graph foundation models on homogeneous graphs, where pre-training and prompt-based adaptation have become two representative paradigms. Early studies such as GPPT~\cite{GPPT} and GraphPrompt~\cite{GraphPrompt} show that graph prompts can effectively bridge pre-training and downstream adaptation, thereby improving transferability under limited supervision. Building on this line, more recent works, such as GCOPE~\cite{GCOPE} and SAMGPT~\cite{SAMGPT}, further explore cross-graph or cross-domain generalization, demonstrating the promise of foundation-style learning beyond a single graph domain. Nevertheless, these methods are primarily designed for homogeneous graphs, where node and edge semantics are relatively unified. As a result, they cannot be directly extended to heterogeneous graphs, whose semantic complexity arises from diverse node types, relation types, and interaction patterns.

\subsection{Foundation Models for Heterogeneous Graphs}
To better capture heterogeneous semantics, recent studies have begun to investigate foundation models tailored for heterogeneous graphs. Representative methods such as HePa~\cite{hepa} and HetGPT~\cite{hetgpt} introduce heterogeneous-aware pre-training strategies, relation-sensitive objectives, or schema-aware modeling mechanisms to learn transferable representations from complex multi-type graphs. In addition, HGPrompt~\cite{hgprompt} further shows that prompt-based adaptation can also be extended to heterogeneous settings, enabling more flexible downstream transfer. However, most of these approaches are still developed under fixed-schema or in-domain settings, where the relation types and semantic structures remain largely stable across training and testing. Consequently, although they are more expressive than homogeneous graph foundation models, they are not well suited for cross-domain heterogeneous transfer, where graph schemas, feature spaces, and structural semantics may differ substantially across domains.

\subsection{LLM-enhanced Heterogeneous Graph Foundation Models}
Another emerging direction incorporates large language models into heterogeneous graph learning. Methods such as HiGPT~\cite{higpt} and LMCH~\cite{LMCH} leverage textual descriptions and external semantic knowledge to align heterogeneous graphs from different domains into a shared semantic space. This strategy provides a promising way to mitigate schema mismatch and improve cross-domain transferability. Nevertheless, the effectiveness of these models fundamentally depends on the availability of rich textual attributes. In many practical applications, such as financial transaction graphs, network traffic graphs, and sensor interaction graphs, node and edge attributes are numerical, categorical, or anonymized rather than textual. In such text-free settings, LLM-based semantic alignment becomes inapplicable. Therefore, existing LLM-enhanced heterogeneous graph foundation models still leave a critical gap in text-free cross-domain heterogeneous learning.

Existing studies have advanced graph foundation models along three major directions, namely homogeneous graph foundation models, heterogeneous graph foundation models, and LLM-enhanced heterogeneous graph foundation models. However, homogeneous approaches fail to preserve heterogeneous semantics, heterogeneous approaches are often limited to fixed-schema settings, and LLM-based approaches rely heavily on textual information. Therefore, a unified framework for text-free, few-shot, cross-domain learning on heterogeneous graphs remains unaddressed. This gap motivates \model, which preserves and transfers heterogeneous structural semantics without relying on external textual supervision.
\section{Preliminaries}

\subsection{Heterogeneous and Homogeneous Graphs}
Heterogeneous graphs possess multiple types of nodes and edges, enabling them to model complex interactive systems in the real world, which distinguishes them from traditional homogeneous graphs that contain only a single node and edge type. Formally, we define a heterogeneous graph as $\mathcal{G}=(\mathcal{V},\mathcal{E},\mathcal{A},\mathcal{R},\phi,\varphi,\mathcal{X}_{\mathcal{V}},\mathcal{X}_{\mathcal{E}})$, where $\mathcal{V}$ and $\mathcal{E}$ denote the sets of nodes and edges, respectively; $\mathcal{A}$ and $\mathcal{R}$ represent the sets of node types and edge types. Each node $v\in\mathcal{V}$ is associated with a specific node type via a mapping function $\phi:\mathcal{V}\rightarrow\mathcal{A}$. Similarly, each edge $e\in\mathcal{E}$ is associated with an edge type via $\varphi:\mathcal{E}\rightarrow\mathcal{R}$. A key characteristic of a heterogeneous graph is that $|\mathcal{A}|+|\mathcal{R}|>2$, meaning there are at least two types of nodes or edges. We denote the node feature set as $\mathcal{X}_{\mathcal{V}}=\{\boldsymbol{X}_{a}\in\mathbb{R}^{|\mathcal{V}_{a}|\times d_{a}}\mid a\in\mathcal{A}\}$, where $\mathcal{V}_a = \{v \in \mathcal{V} \mid \phi(v) = a\}$ is the set of nodes of type $a$, $\boldsymbol{X}_{a}$ is the original feature matrix for these nodes, and $d_{a}$ is the feature dimension for type $a$. The edge feature set is $\mathcal{X}_{\mathcal{E}}=\{\boldsymbol{X}_{r}\in\mathbb{R}^{|\mathcal{E}_{r}\mid\times d_{r}}\mid r\in\mathcal{R}\}$, where $\boldsymbol{X}_{r}$ represents the feature matrix for edges of type $r$.

To address the structural mismatch issues caused by heterogeneity, we can transform the heterogeneous graph into a homogeneous counterpart. In this context, the homogeneous graph not only implies uniform node and edge types but also dictates that all nodes and edges have been projected into a unified latent feature space. We define the transformed homogeneous graph as $G=(V,E,\boldsymbol{A},\boldsymbol{X}_{V},\boldsymbol{X}_{E})$. Here, $V$ and $E$ denote the transformed node and edge sets, respectively, and $\boldsymbol{A}$ is the adjacency matrix describing the graph connectivity. To resolve the feature dimension discrepancy ($d_a$) across different domains and node types, we employ Singular Value Decomposition (SVD) to reduce and align the original features, obtaining the aligned node feature matrix $\boldsymbol{X}_{V}\in\mathbb{R}^{|V|\times d}$. The aligned edge feature matrix is denoted as $\boldsymbol{X}_{E}\in\mathbb{R}^{|E|\times f}$. Importantly, we preserve the semantic information from the original heterogeneous graph in the form of edges; thus, rich semantic features are encapsulated within the edges of the transformed homogeneous graph.

\subsection{Graph Encoder}
We adopt a Graph Neural Network (GNN) based on the message-passing mechanism as our backbone encoder. Specifically, we utilize a universal graph encoder $f_{\theta}$, parameterized by $\boldsymbol{\theta}$, to learn high-order representations of nodes. The computation entails stacking multiple message-passing layers. Let $\boldsymbol{h}_{v}^{(l)}$ denote the hidden embedding of node $v$ at the $l$-th layer. The update process is formulated as:
\begin{equation}
\boldsymbol{h}_{v}^{(l)}=\text{Aggr}\left(\boldsymbol{h}_{v}^{(l-1)},\left\{\left(\boldsymbol{h}_{u}^{(l-1)},\boldsymbol{e}_{uv}\right):u\in\mathcal{N}_{v}\right\};\boldsymbol{\theta}^{(l)}\right)
\end{equation}
where $\mathcal{N}_{v}$ denotes the set of neighbors for node $v$, and $\boldsymbol{e}_{uv}$ represents the semantic edge feature connecting nodes $u$ and $v$ (which inherently corresponds to a row vector in $\boldsymbol{X}_{E}$). This edge feature explicitly participates in the aggregation process, ensuring that the model leverages the original heterogeneous information preserved on the edges. The function $\text{Aggr}(\cdot;\boldsymbol{\theta}^{(l)})$ represents the message-passing and aggregation function at the $l$-th layer, governed by the learnable parameters $\boldsymbol{\theta}^{(l)}\in\boldsymbol{\theta}$.

\subsection{Target Domain and Few-shot Classification Task}
In cross-domain transfer learning, we assume the model is pre-trained on multiple source domains containing abundant unlabeled data. Following pre-training, our objective is to transfer the extracted universal knowledge to a target graph $G_{T}$ with a different data distribution.

To enhance framework universality, we model downstream node classification and graph classification tasks as \textit{instance-level classification}. Specifically, an ``instance'' can refer to a single node in the target graph $G_{T}$ or a local subgraph obtained through multi-hop sampling centered on a specific node.

For the few-shot instance classification task on the target domain, we assume the target dataset comprises $|C|$ distinct instance categories, forming a label space $\mathcal{Y}=\{1,2,...,|C|\}$. Under the stringent few-shot constraint, we can only access a small number of labeled instances, $K$, for each category $m\in\mathcal{Y}$ in the target domain. We define this collection of $|C|\times K$ labeled samples as the training set $\mathcal{S}$ for the few-shot fine-tuning phase. The ultimate task of the model is to adapt the prior prompts using this extremely limited labeled set $\mathcal{S}$, and to accurately predict the categories of the remaining unknown instances in the target dataset, which are strictly partitioned into a validation set and a test set.

\begin{figure*}[htbp]
    \centering
    \includegraphics[width=\linewidth]{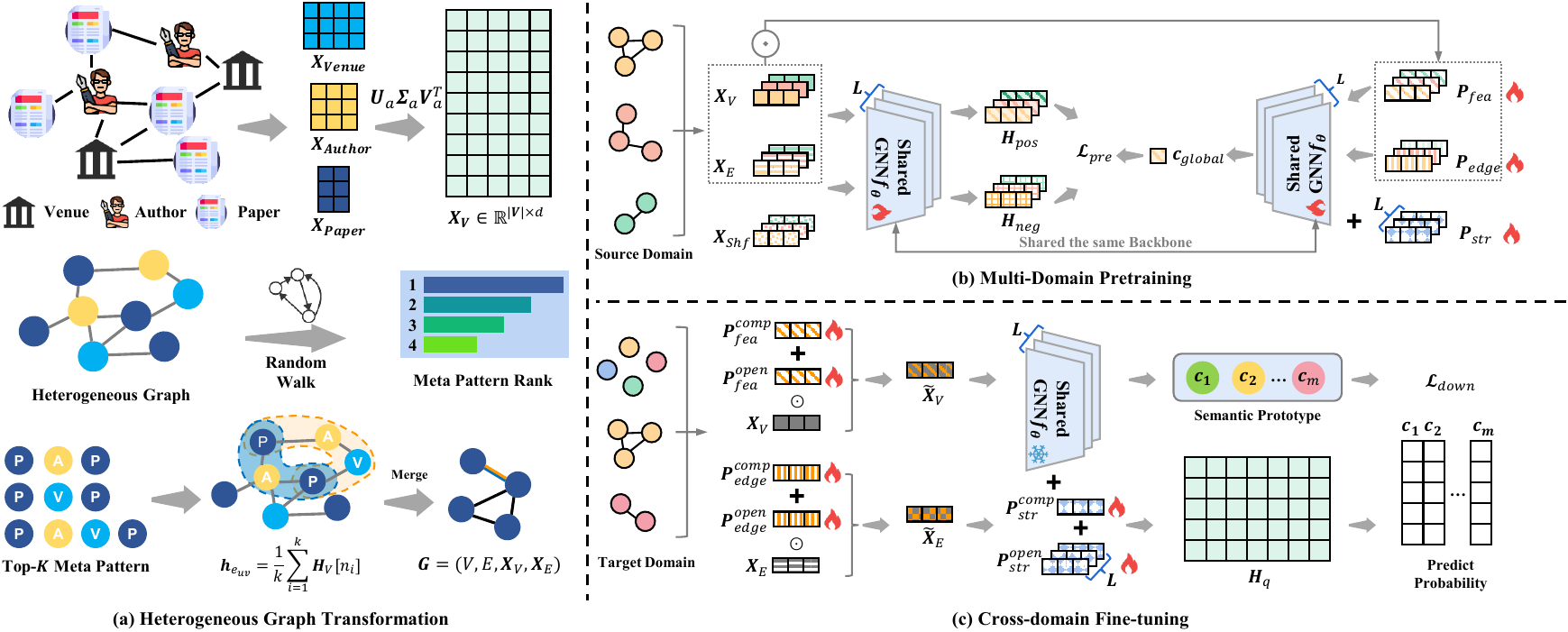}
    \caption{Overview of the \model framework. The workflow consists of three phases:
    1) \textbf{Semantic-preserving graph transformation}: each source and target heterogeneous graph is converted into a unified homogeneous graph through SVD-based feature alignment and meta-pattern mining, while heterogeneous semantics are preserved in semantics-enriched edge features;
    2) \textbf{Multi-domain semantic graph pre-training}: a Tri-Prompt mechanism with feature, edge, and structure prompts is integrated with a shared GNN backbone to capture transferable knowledge from feature, relation, and topology perspectives through self-supervised contrastive learning;
    3) \textbf{Cross-domain semantic fine-tuning}: the pre-trained backbone is frozen, source-domain prompts are composed with target-specific open prompts for few-shot adaptation, and the adapted representations are used for prototype-based prediction on target-domain instances.}
    \label{fig:framework}
\end{figure*}

\section{Method}
This section presents the technical details of \model, which consists of three phases: semantic-preserving graph transformation, multi-domain semantic graph pre-training, and cross-domain semantic fine-tuning. First, heterogeneous graphs are converted into unified homogeneous graphs through SVD-based feature alignment and data-driven meta-pattern mining, with heterogeneous interactions preserved in semantics-enriched edges. Next, a Tri-Prompt mechanism, including feature, edge, and structure prompts, is integrated with a shared GNN backbone for self-supervised multi-domain pre-training under the Deep Graph Infomax (DGI) paradigm. Finally, for the target domain, the pre-trained backbone is frozen and lightweight prompt modules are adapted via prompt composition and prototype-based few-shot learning. The overall architecture is illustrated in Fig.~\ref{fig:framework}, and the complete cross-domain learning procedure is summarized in Algorithm~\ref{alg:crosshgl}.

\subsection{Semantic-Preserving Graph Transformation}\label{sec: SPGT}
To address the issues of inconsistent node feature dimensions in heterogeneous graphs and the difficulty of directly adapting heterogeneous structures to a unified graph model, a semantic-preserving graph transformation strategy is proposed. This strategy maps a heterogeneous graph into a feature-aligned homogeneous graph $G$ with semantics-enriched edges. The specific process involves the following three stages:

\subsubsection{Feature Space Alignment}
Nodes of different types in heterogeneous graphs typically possess varying feature dimensions, and some nodes may even lack features. To eliminate this feature space heterogeneity, Singular Value Decomposition (SVD) is first employed to reduce the dimensionality and align the features across all node types.
For each node type $a\in\mathcal{A}$, given its raw feature matrix $\boldsymbol{X}_{a}\in\mathbb{R}^{|\mathcal{V}_{a}|\times d_{a}}$, SVD is performed as follows:
\begin{equation}
\boldsymbol{X}_{a}\approx \boldsymbol{U}_{a}\boldsymbol{\Sigma}_{a}\boldsymbol{V}_{a}^{\top}
\end{equation}
Specifically, a unified target feature dimension $d$ is predefined. To derive the dense representations, the matrices $\boldsymbol{U}_{a}$ and $\boldsymbol{\Sigma}_{a}$ are truncated to retain only the top-$d$ singular values and their corresponding left singular vectors, denoted as $\boldsymbol{U}_{a,d}\in\mathbb{R}^{|\mathcal{V}_{a}|\times d}$ and $\boldsymbol{\Sigma}_{a,d}\in\mathbb{R}^{d\times d}$, respectively. The dimension-reduced feature matrix $\tilde{\boldsymbol{X}}_{a}$ for node type $a$ is then computed by multiplying the truncated left singular matrix with the diagonal matrix of the top-$d$ singular values:
\begin{equation}
\tilde{\boldsymbol{X}}_{a}=\boldsymbol{U}_{a,d}\boldsymbol{\Sigma}_{a,d}
\end{equation}

\begin{algorithm}[!htb]
\caption{Cross-domain Learning Procedure of \model}
\label{alg:crosshgl}
\begin{algorithmic}[1]
\STATE {\bfseries Input:} Source heterogeneous graphs $\{\mathcal{G}^{(k)}\}_{k=1}^{K}$, target heterogeneous graph $\mathcal{G}_T$, target support set $\mathcal{S}$.
\STATE {\bfseries Output:} Predictions for target-domain instances.

\STATE \textbf{Initialization:} Initialize a shared GNN backbone $f_{\boldsymbol{\theta}}$ and source-domain Tri-Prompts $\{\boldsymbol{P}_{\text{fea}}^{(k)}, \boldsymbol{P}_{\text{edge}}^{(k)}, \boldsymbol{P}_{\text{str}}^{(k)}\}_{k=1}^{K}$.

\STATE \textcolor{gray}{// Phase 1: Semantic-preserving graph transformation}
\FOR{each graph $\mathcal{G} \in \{\mathcal{G}^{(k)}\}_{k=1}^{K} \cup \{\mathcal{G}_T\}$}
    \STATE Align type-specific node features $\mathcal{X}_{\mathcal{V}}$ into a unified feature space $\boldsymbol{X}_V$ via SVD.
    \STATE Mine the Top-$K$ meta-pattern set $\mathcal{P}=\{P_1,\dots,P_m\}$ by constrained random walks.
    \STATE Construct the transformed homogeneous graph: $G=(V,E,\boldsymbol{A},\boldsymbol{X}_V,\boldsymbol{X}_E)$.
\ENDFOR

\STATE \textcolor{gray}{// Phase 2: Multi-domain semantic graph pre-training}
\FOR{each source domain $k=1$ to $K$}
    \STATE Apply $\boldsymbol{P}_{\text{all}}^{(k)}=\{\boldsymbol{P}_{\text{fea}}^{(k)},\boldsymbol{P}_{\text{edge}}^{(k)},\boldsymbol{P}_{\text{str}}^{(k)}\}$ to the transformed source graph.
    \STATE Construct positive, augmented, and negative views: $(\boldsymbol{X}_V,\boldsymbol{A},\boldsymbol{X}_E)$, $(\tilde{\boldsymbol{X}}_V,\tilde{\boldsymbol{A}},\tilde{\boldsymbol{X}}_E)$, and $(\boldsymbol{X}_{\text{Shf}},\boldsymbol{A},\boldsymbol{X}_E)$.
    \STATE Obtain node representations $\boldsymbol{H}_{\text{pos}}$, $\boldsymbol{H}_{\text{neg}}$, and graph summary $\boldsymbol{c}_{\text{global}}$ using $f_{\boldsymbol{\theta}}$.
    \STATE Update $f_{\boldsymbol{\theta}}$, $\boldsymbol{P}_{\text{all}}^{(k)}$ by minimizing contrastive loss $\mathcal{L}_{\text{pre}}$.
\ENDFOR

\STATE \textcolor{gray}{// Phase 3: Cross-domain semantic fine-tuning}
\STATE Freeze the pre-trained backbone $f_{\boldsymbol{\theta}}$.
\STATE Construct composed prompts $\boldsymbol{P}_*^{\text{comp}}$ and initialize target-specific open prompts $\boldsymbol{P}_*^{\text{open}}$, where $* \in \{\text{fea}, \text{edge}, \text{str}\}$.
\STATE Adapt target prompts on $\mathcal{G}_T$ with the few-shot support set $\mathcal{S}$ and obtain target representations $\boldsymbol{Z}_{\text{final}}$.
\STATE Compute class prototypes $\boldsymbol{c}_m$ from the adapted target representations.
\STATE Predict labels for target-domain instances via prototype matching $p(y=m \mid \boldsymbol{h}_q)$.

\STATE \textbf{return} Target-domain predictions.
\end{algorithmic}
\end{algorithm}

Prior to the SVD application, a pre-processing step is executed to address dimensional deficiencies. For node types completely lacking initial attributes, one-hot vectors are employed for feature initialization. Additionally, in cases where the original feature dimension falls short of the target dimension (i.e., $d_{a}<d$), zero-padding is applied directly to the raw feature matrix $\boldsymbol{X}_{a}$ to explicitly extend its dimensionality to $d$. Ultimately, by unifying these pre-processed features through the aforementioned SVD and truncation, a global node feature matrix $\boldsymbol{X}_{V}\in\mathbb{R}^{|V|\times d}$ is constructed, ensuring strict dimensional consistency across all nodes in the graph.
\subsubsection{Data-Driven Mining of Meta-Patterns}
Unlike traditional methods that rely on expert knowledge to predefine meta-paths, we adopt a data-driven approach to automatically mine high-frequency meta-patterns for the target node type.
Specifically, we execute constrained random walks on the heterogeneous graph. For each target-type node, walk sequences of length $L$ are generated. Subsequently, a sliding window of size $w$ is applied to these sequences to systematically extract local sub-paths. To capture interaction patterns with closed-loop semantics, node type sequences within these windows where both the starting and ending node types correspond to $v_{\text{target}}$ are filtered out as candidate meta-patterns. These candidates are denoted as $\mathcal{P}=\{P_{1}, P_{2},\dots, P_{m}\}$, where $P_{i}$ represents a specific sequence of node types (e.g., Paper-Author-Paper).
Finally, the frequencies of all candidate patterns are counted, and the Top-$K$ key meta-patterns are selected based on their occurrence. This process not only discovers explicit relationships but also captures latent semantic dependencies.

\subsubsection{Semantic Graph Homogenization}
Guided by the mined meta-pattern set $\mathcal{P}$, the target homogeneous graph $G=(V,E,\boldsymbol{A},\boldsymbol{X}_{V},\boldsymbol{X}_{E})$ is constructed. Specifically, the node set $V$ exclusively retains target-type nodes, defined as $V=\{v\in\mathcal{V} \mid \phi(v)=v_{\text{target}}\}$, with aligned features $\boldsymbol{X}_{V}$. For each meta-pattern $P_i\in\mathcal{P}$, a new edge $(u,v)$ is established in $E$ whenever a matching path instance connects two distinct target nodes ($u,v\in V$, $u\ne v$) in the original heterogeneous graph. 

In densely connected heterogeneous networks, multiple path instances often concurrently connect the same pair of target nodes. Let $\mathcal{I}_{uv}$ denote the set of all valid path instances connecting nodes $u$ and $v$ under the pattern set $\mathcal{P}$. To comprehensively preserve the rich semantic context during this topological transformation, the features of all intermediate nodes across all path instances in $\mathcal{I}_{uv}$ are simultaneously aggregated to formulate the semantic edge feature vector $\boldsymbol{e}_{uv}$:
\begin{equation}
\boldsymbol{e}_{uv} = \frac{1}{|\mathcal{I}_{uv}|} \sum_{p\in\mathcal{I}_{uv}} \left( \frac{1}{k_p} \sum_{j=1}^{k_p} \boldsymbol{X}_{V}[n_{j}^{(p)}] \right)
\end{equation}
where $k_p$ represents the number of intermediate nodes in a specific path instance $p$, and $n_{j}^{(p)}$ denotes the $j$-th intermediate node within that path. Subsequently, the collection of all aggregated edge vectors $\boldsymbol{e}_{uv}$ constitutes the semantics-enriched edge feature matrix $\boldsymbol{X}_{E}$ for the target homogeneous graph.

Through the aforementioned process, the complex heterogeneous graph topology is transformed into a homogeneous graph containing only a single node type but highly enriched edge features. This enables a universal homogeneous graph encoder to efficiently process the graph structure while simultaneously perceiving the complex semantic interactions of the original heterogeneous graph.

\subsection{Multi-Domain Semantic Graph Pre-training}
To enable Graph Neural Networks (GNNs) to learn universal graph representations in an unsupervised setting and effectively utilize the previously constructed homogeneous graphs rich in meta-pattern semantics, a Tri-Prompt Graph Contrastive Pre-training Framework is proposed.
The core idea of this framework is to decouple the adaptation processes of node features, graph topologies, and edge semantics. By utilizing lightweight prompts to guide a shared GNN backbone, the model captures universal knowledge from diverse perspectives and optimizes it via Local-Global Mutual Information Maximization in a self-supervised manner.

\subsubsection{Tri-Prompt Design}
As detailed in Section~\ref{sec: SPGT}, the original heterogeneous graph is uniformly transformed into a target-node homogeneous graph $G=(V,E,\boldsymbol{A},\boldsymbol{X}_{V},\boldsymbol{X}_{E})$. Here, $\boldsymbol{X}_{V}$ encapsulates the aligned node features, while $\boldsymbol{X}_{E}$ acts as a rich semantic container for the intermediate heterogeneous interactions. To fully exploit and seamlessly decouple this multi-dimensional information without altering the pre-trained backbone, the Tri-Prompt module is designed. From a functional perspective, this framework operates sequentially across two highly synergistic phases: input-level semantic filtration and layer-level topological intervention.

During the initial phase of \textit{input-level semantic filtration}, both node features and edge semantics often carry universal knowledge that may contain task-irrelevant noise for a specific downstream domain. To address this, a unified element-wise modulation strategy is employed to adaptively filter and align the input spaces. Specifically, the feature prompt $\boldsymbol{P}_{\text{fea}}$ and the semantic edge prompt $\boldsymbol{P}_{\text{edge}}$ act as learnable dynamic filters that smoothly project the initial features into a task-specific latent space prior to deep encoding:
\begin{equation}
\tilde{\boldsymbol{X}}_{V} = \boldsymbol{X}_{V} \odot \boldsymbol{P}_{\text{fea}}, \quad \tilde{\boldsymbol{X}}_{E} = \boldsymbol{X}_{E} \odot \boldsymbol{P}_{\text{edge}}
\end{equation}
While the modulated node feature matrix $\tilde{\boldsymbol{X}}_{V}$ directly serves as the initial hidden states for the GNN, the purified edge feature matrix $\tilde{\boldsymbol{X}}_{E}$ requires further mapping to guide structural routing. Because these edge features deeply compress heterogeneous meta-paths, they are subsequently transformed into scalar attention weights $\alpha_{uv}$ via a dedicated edge encoder:
\begin{equation}
\alpha_{uv}=\sigma(\text{MLP}_{\text{edge}}(\tilde{\boldsymbol{e}}_{uv})), \quad \forall(u,v)\in E
\end{equation}
By unifying the feature and edge prompts at the entry point of the network, this filtration phase ensures that the model selectively activates only the node attributes and interaction paths most critical to the current domain, effectively stripping away redundant background noise.

Subsequent to the input space calibration, the framework executes a \textit{layer-level topological intervention} driven by the structure prompt $\boldsymbol{P}_{\text{str}}$. Operating directly on the hidden layers of the GNN, this prompt deeply intervenes in the topology-aware mechanisms of message passing. The encoder modulated by the structure prompt is denoted as $f_{\boldsymbol{\theta}}^{\prime}$. For the hidden state matrix $\boldsymbol{H}^{(l)}$ produced by the $l$-th GNN layer, the layer-specific structure prompt matrix $\boldsymbol{P}_{\text{str}}^{(l)}$ dynamically updates the node representations through residual modulation:
\begin{equation}
\boldsymbol{Z}^{(l)}=\text{Aggregate}(\boldsymbol{H}^{(l)},\boldsymbol{A},\boldsymbol{\alpha})
\end{equation}
\begin{equation}
\boldsymbol{H}^{(l+1)}=\text{ReLU}((\boldsymbol{Z}^{(l)}+\boldsymbol{H}^{(l)})\odot \boldsymbol{P}_{\text{str}}^{(l)})
\end{equation}
Through the integration of these three orthogonal prompts, the framework achieves a highly expressive yet parameter-efficient adaptation, elegantly decoupling the initial semantic alignment from the deep structural aggregation.

\subsubsection{Prompt-Aware Graph Contrastive Pre-training}
Having established the tri-prompt system, the Deep Graph Infomax (DGI) paradigm is adopted to execute unified self-supervised pre-training. The entire pre-training pipeline fundamentally consists of three sequential stages: view construction, dual-path encoding, and discriminator optimization.

To construct the positive and negative sample pairs required for contrastive learning, specific augmentation strategies~\cite{GDAug} are applied in the topological and feature spaces. The positive view strictly retains the local structure and features of the original graph, denoted as the tuple $(\boldsymbol{X}_{V},\boldsymbol{A},\boldsymbol{X}_{E})$. Concurrently, an augmented view is generated by perturbing the topology through random edge dropping with probability $p$, yielding an augmented adjacency matrix $\tilde{\boldsymbol{A}}$ and a corresponding edge feature subset $\tilde{\boldsymbol{X}}_{E}$. To establish a contrastive baseline, a negative view is constructed by randomly shuffling the node features along the node dimension. This shuffling operation deliberately breaks the joint correspondence between the graph topology and node attributes to introduce spurious node semantics, yielding the shuffled feature matrix:
\begin{equation}
\boldsymbol{X}_{\text{Shf}} = \text{Shuffle}(\boldsymbol{X}_{V})
\end{equation}
The negative view thereby preserves the original topology $\boldsymbol{A}$ alongside these corrupted features.

During the encoding phase, the shared GNN backbone $f_{\boldsymbol{\theta}}$, coupled with distinct prompt pathways, extracts three levels of graph representations corresponding to the constructed views. The positive anchor view is fed into the base backbone to acquire the genuine local context features $\boldsymbol{H}_{\text{pos}}$ for all nodes:
\begin{equation}
\boldsymbol{H}_{\text{pos}} = f_{\boldsymbol{\theta}}(\boldsymbol{X}_{V},\boldsymbol{A},\boldsymbol{X}_{E})
\end{equation}
Similarly, the negative view is input into the backbone to obtain the counterfeit node representation matrix $\boldsymbol{H}_{\text{neg}}$:
\begin{equation}
\boldsymbol{H}_{\text{neg}} = f_{\boldsymbol{\theta}}(\boldsymbol{X}_{\text{Shf}},\boldsymbol{A},\boldsymbol{X}_{E})
\end{equation}
For the global graph summary, the augmented view is processed by the backbone equipped with the complete tri-prompt set, denoted collectively as $\boldsymbol{P}_{\text{all}} = \{\boldsymbol{P}_{\text{fea}},\boldsymbol{P}_{\text{edge}},\boldsymbol{P}_{\text{str}}\}$. A readout function $\mathcal{R}$ subsequently aggregates all perturbed node representations to generate the graph-level global summary:
\begin{equation}
\boldsymbol{c}_{\text{global}} = \sigma\left(\frac{1}{|V|}\sum_{i\in V}\left[f_{\boldsymbol{\theta}}\left(\tilde{\boldsymbol{X}}_{V},\tilde{\boldsymbol{A}},\tilde{\boldsymbol{X}}_{E},\boldsymbol{P}_{\text{all}}\right)\right]_{i}\right)
\end{equation}

The pre-training optimization objective is to enable the global graph summary $\boldsymbol{c}_{\text{global}}$ to accurately identify its genuine constituents (positive node vectors $\boldsymbol{h}_{\text{pos}}$) and repel the misaligned counterfeit nodes (negative node vectors $\boldsymbol{h}_{\text{neg}}$). To this end, a bilinear discriminator $\mathcal{D}$, parameterized by a weight matrix $\boldsymbol{W}$, is introduced to calculate the similarity scores between node-level representations and the global summary:
\begin{equation}
\mathcal{D}(\boldsymbol{h},\boldsymbol{c}) = \sigma(\boldsymbol{h}^\top \boldsymbol{W} \boldsymbol{c})
\end{equation}
The final pre-training loss function $\mathcal{L}_{\text{pre}}$ is defined as a standard Binary Cross-Entropy (BCE) loss:
\begin{equation}
\resizebox{\columnwidth}{!}{$
\displaystyle
\mathcal{L}_{\text{pre}} = -\frac{1}{|V|}\sum_{i=1}^{|V|}[\log\mathcal{D}(\boldsymbol{h}_{\text{pos},i},\boldsymbol{c}_{\text{global}}) + \log(1-\mathcal{D}(\boldsymbol{h}_{\text{neg},i},\boldsymbol{c}_{\text{global}}))]
$}
\end{equation}
Throughout the multi-domain pre-training phase, the model shares the same GNN backbone parameters across $K$ different source domain datasets, but independently optimizes a specific set of tri-prompts $\{\boldsymbol{P}_{\text{fea}}^{(k)},\boldsymbol{P}_{\text{edge}}^{(k)},\boldsymbol{P}_{\text{str}}^{(k)}\}_{k=1}^{K}$ for each dataset.

\subsection{Cross-domain Semantic Fine-tuning}
Upon completion of the multi-domain pre-training, the shared GNN backbone $f_{\boldsymbol{\theta}}$ learns transferable graph representations, while a prior prompt knowledge base $\{\boldsymbol{P}_{\text{fea}}^{(k)},\boldsymbol{P}_{\text{edge}}^{(k)},\boldsymbol{P}_{\text{str}}^{(k)}\}_{k=1}^{K}$ is obtained from the source domains. For $N$-way $K$-shot cross-domain adaptation on the target domain, the backbone parameters are kept frozen, and a Tri-Prompt Composition and Decoupled Fusion Mechanism is introduced to adapt prior knowledge to the new domain.

\subsubsection{Tri-Prompt Composition Mechanism}
For prompt modules across all dimensions, a joint construction strategy combining a composed prompt and an open prompt is adopted to balance prior knowledge transfer with target domain adaptation. Specifically, to dynamically distill the universal knowledge from the $K$ source domains, a learnable attention vector $\boldsymbol{\lambda}\in\mathbb{R}^{K}$ is introduced. The normalized attention weight $w_k$ for the $k$-th pre-trained prompt is computed via Softmax, yielding the composed prompt $\boldsymbol{P}_{*}^{\text{comp}}$ for any specific prompt type $*\in\{\text{fea}, \text{edge}, \text{str}\}$ through a weighted summation:
\begin{equation}
w_k^{*} = \frac{\exp(\lambda_k^{*})}{\sum_{j=1}^K \exp(\lambda_j^{*})}, \quad \boldsymbol{P}_{*}^{\text{comp}} = \sum_{k=1}^K w_k^{*}\boldsymbol{P}_{*}^{(k)}
\end{equation}
Simultaneously, a task-specific open prompt $\boldsymbol{P}_{*}^{\text{open}}$ is randomly initialized to absorb the specific distribution of the new target domain.
During the forward pass of fine-tuning, the edge semantics are first adaptively filtered. The updated edge feature matrix $\tilde{\boldsymbol{X}}_{E}$ is jointly modulated by the composed edge prompt and the open edge prompt:
\begin{equation}
\tilde{\boldsymbol{X}}_{E}=\boldsymbol{P}_{\text{edge}}^{\text{comp}}(\boldsymbol{X}_{E})+\beta\cdot \boldsymbol{P}_{\text{edge}}^{\text{open}}(\boldsymbol{X}_{E})
\end{equation}
where $\beta$ is a learnable coefficient controlling the intensity of edge semantic fusion in the target domain.

Following the edge semantic modulation, the node representations are inferred through a decoupled dual-branch architecture. In the feature-focused branch, the initial node features $\boldsymbol{X}_{V}$ are mapped using the feature prompt, and combined with the purified semantic edge features $\tilde{\boldsymbol{X}}_{E}$ as inputs to the frozen backbone $f_{\boldsymbol{\theta}}$. This branch solely reshapes the external feature space to extract the feature-driven embedding matrix $\boldsymbol{Z}_{\text{fea}}$, without injecting any additional structure prompts:
\begin{equation}
\tilde{\boldsymbol{X}}_{V}=\boldsymbol{P}_{\text{fea}}^{\text{comp}}(\boldsymbol{X}_{V})+\gamma_{\text{fea}}\cdot \boldsymbol{P}_{\text{fea}}^{\text{open}}(\boldsymbol{X}_{V})
\end{equation}
\begin{equation}
\boldsymbol{Z}_{\text{fea}}=f_{\boldsymbol{\theta}}(\tilde{\boldsymbol{X}}_{V},\boldsymbol{A},\tilde{\boldsymbol{X}}_{E})
\end{equation}

Parallel to the feature branch, the structure-focused branch aims to adjust the message passing preferences under the target domain's topology. Here, the original node features $\boldsymbol{X}_{V}$ are kept unchanged, and the structure prompt $\boldsymbol{P}_{\text{str}}$ is directly injected into the frozen GNN. The final structure-driven embedding matrix $\boldsymbol{Z}_{\text{str}}$ is inferred jointly by the prior composed structure prompt and the open structure prompt:
\begin{equation}
\boldsymbol{Z}_{\text{str}}=f_{\boldsymbol{\theta}}(\boldsymbol{X}_{V},\boldsymbol{A},\tilde{\boldsymbol{X}}_{E};\boldsymbol{P}_{\text{str}}^{\text{comp}})+\gamma_{\text{str}}\cdot f_{\boldsymbol{\theta}}(\boldsymbol{X}_{V},\boldsymbol{A},\tilde{\boldsymbol{X}}_{E};\boldsymbol{P}_{\text{str}}^{\text{open}})
\end{equation}
where $\gamma_{\text{str}}$ dynamically balances the prior topological rules with the novel topological rules of the target domain. Finally, the comprehensive representations $\boldsymbol{Z}_{\text{final}}$ is obtained through the weighted residual fusion of both branches:
\begin{equation}
\boldsymbol{Z}_{\text{final}}=\boldsymbol{Z}_{\text{fea}}+\alpha\cdot \boldsymbol{Z}_{\text{str}}
\end{equation}
where $\alpha$ is the branch balancing coefficient.
\subsubsection{Prototype-based Classification and Robust Evaluation}
To enhance the universality of the framework, the downstream tasks are uniformly modeled as instance-level classification problems, employing non-parametric Prototypical Networks to mitigate overfitting in few-shot scenarios. After obtaining the final node representations $\boldsymbol{Z}_{\text{final}}$, a unified instance-level embedding vector $\boldsymbol{h}_{i}$ is constructed for each target sample $i$. For node classification tasks, the instance representation is directly derived from the final node embedding, such that $\boldsymbol{h}_{i}=\boldsymbol{Z}_{\text{final}}[i]$. Conversely, for graph or subgraph classification tasks, a local subgraph $\mathcal{G}_{i}$ is extracted via multi-hop sampling centered on the target node, and a readout function (e.g., mean pooling) aggregates the node representations within the subgraph to yield the global representation $\boldsymbol{h}_{i}=\text{Readout}(\{\boldsymbol{Z}_{\text{final}}[j] \mid j\in \mathcal{G}_{i}\})$.

During the few-shot training phase, a support set $\mathcal{S}$ is constructed by randomly sampling $K$ labeled instances for each category $m$. The semantic prototype vector $\boldsymbol{c}_{m}$ for category $m$ is then computed by averaging the embeddings of all instances belonging to the same class within $\mathcal{S}$:
\begin{equation}
\boldsymbol{c}_{m}=\frac{1}{K}\sum_{i\in\mathcal{S}_{m}}\boldsymbol{h}_{i}
\end{equation}
For any unknown query instance $\boldsymbol{h}_{q}$ in the test set, the cosine similarity between its normalized embedding and each category prototype $\boldsymbol{c}_{m}$ is calculated. This similarity is converted into a prediction probability via the Softmax function:
\begin{equation}
p(y=m|\boldsymbol{h}_{q})=\frac{\exp(\text{sim}(\boldsymbol{h}_{q},\boldsymbol{c}_{m})/\tau)}{\sum_{j=1}^{|C|}\exp(\text{sim}(\boldsymbol{h}_{q},\boldsymbol{c}_{j})/\tau)}
\end{equation}
The optimization objective for fine-tuning minimizes the cross-entropy loss on the training set. Since the backbone network is completely frozen, only the lightweight prompt parameters are updated. To eliminate the random bias introduced by few-shot sampling, an early stopping is employed on a validation set, and fine-tuning is independently repeated multiple times using different random seeds to ensure robust evaluation.

\section{Evaluations}\label{sec: experiment}
\subsection{Experimental Settings}

\subsubsection{Datasets}
Extensive experiments are conducted on four widely used real-world heterogeneous graph benchmarks for cross-domain evaluation: ACM, DBLP~\cite{lv2021we}, IMDB~\cite{IMDB}, and Freebase~\cite{lv2021we}. During the feature initialization phase, any raw textual attributes associated with the nodes are directly pre-processed into pure numerical vectors. By intentionally stripping away all natural language semantics and precluding the use of LLM-based alignment, this setup forces the models to rely strictly on the inherent topological structures and relational meta-patterns for cross-domain transfer.

Table~\ref{tab:dataset_summary} summarizes the basic statistics of the four heterogeneous graph benchmarks. These datasets differ not only in graph scale, but also in schema complexity, as reflected by the numbers of node types and edge types. Such diversity provides a suitable testbed for evaluating the cross-domain transferability of heterogeneous graph foundation models.

\begin{table}[!htb]
  \centering
  \caption{Summary of the heterogeneous graph datasets used in our cross-domain evaluation.}
  \label{tab:dataset_summary}
  \renewcommand\arraystretch{1.3}
  \resizebox{\linewidth}{!}{
    \begin{tabular}{c|c|c|c|c|c}
      \hline\hline
      \multirow{2}{*}{\textbf{Dataset}} & \multirow{2}{*}{\textbf{\#Nodes}} & \textbf{\#Node} & \multirow{2}{*}{\textbf{\#Edges}} & \textbf{\#Edge} & \multirow{2}{*}{\textbf{\#Classes}} \\
       &  & \textbf{Types} &  & \textbf{Types} &  \\
      \hline
      ACM      & 10,942  & 4 & 547,872   & 8  & 3 \\
      DBLP     & 26,128  & 4 & 239,566   & 6  & 4 \\
      IMDB     & 11,616  & 3 & 34,212    & 4  & 3 \\
      Freebase & 180,098 & 8 & 1,057,688 & 36 & 7 \\
      \hline\hline
    \end{tabular}
  }
\end{table}

\subsubsection{Baselines}
To comprehensively demonstrate the superiority of \model, it is evaluated against ten representative state-of-the-art baselines, which are systematically categorized into four groups:
\textbf{(1) Homogeneous GNNs:} GCN~\cite{GCN}, GraphSAGE~\cite{GraphSAGE}, and GAT~\cite{velivckovic2017gat}, which inherently flatten the heterogeneous topology to perform message passing;
\textbf{(2) Heterogeneous GNNs:} HAN~\cite{HAN} and Simple-HGN~\cite{simple-HGN}, which are representative HGNNs that account for heterogeneity through neighborhood aggregation guided by meta-paths or edge-type-specific attention;
\textbf{(3) Graph Contrastive Learning:} GraphCL~\cite{GraphCL} and DGI~\cite{DGI}, which are representative self-supervised pre-training methods designed for universal graph representation learning;
\textbf{(4) Graph Foundation Models:} GPPT~\cite{GPPT}, GraphPrompt~\cite{GraphPrompt}, GCOPE~\cite{GCOPE}, and SAMGPT~\cite{SAMGPT}, which encompass recent explorations in prompt-based graph learning. 

\begin{table*}[!htb]
  \centering
  \caption{Few-shot node classification results: average Micro-F1 and Macro-F1 (\%) $\pm$ standard deviation across four cross-domain tasks. \textbf{Boldface} letters mark the best performance, while \underline{underlined} letters indicate the second best.}
  \renewcommand\arraystretch{1.3}
  \label{tab:main_results}
  \resizebox{\textwidth}{!}{
    \begin{tabular}{cc|c|cc|cc|cc|cc} 
    \hline\hline
    \multirow{2}{*}{\textbf{Paradigm}} & \multirow{2}{*}{\textbf{Category}} & \multirow{2}{*}{\textbf{Method}} & \multicolumn{2}{c|}{ACM} & \multicolumn{2}{c|}{DBLP} & \multicolumn{2}{c|}{IMDB} & \multicolumn{2}{c}{Freebase} \\ 
    \cline{4-11}
    & & & Micro-F1 & Macro-F1 & Micro-F1 & Macro-F1 & Micro-F1 & Macro-F1 & Micro-F1 & Macro-F1 \\ 
    \hline
    \multirow{5}{*}{\begin{tabular}[c]{@{}c@{}}Supervised\\ Learning\end{tabular}} 
    & \multirow{2}{*}{Heterogeneous} & HAN         & 52.66 $\pm$ 3.90 & 49.20 $\pm$ 6.99 & 41.72 $\pm$ 5.52 & 36.68 $\pm$ 6.84 & 37.25 $\pm$ 2.15 & 27.68 $\pm$ 5.49 & \textbf{40.12 $\pm$ 7.39} & 8.01 $\pm$ 1.29 \\
    &                                & Simple-HGN  & 48.15 $\pm$ 5.44 & 45.15 $\pm$ 3.67 & 37.53 $\pm$ 2.16 & 33.26 $\pm$ 4.16 & 35.14 $\pm$ 2.42 & 24.15 $\pm$ 4.31 & \uline{37.73 $\pm$ 4.27} & 7.52 $\pm$ 2.32 \\ 
    \cdashline{3-11}
    & \multirow{3}{*}{Homogeneous}   & GCN         & \uline{61.02 $\pm$ 10.23}& \uline{56.19 $\pm$ 12.92}& 37.95 $\pm$ 4.98 & 34.84 $\pm$ 6.16 & 37.09 $\pm$ 2.06 & 27.20 $\pm$ 4.92 & 22.42 $\pm$ 0.19 & 4.02 $\pm$ 0.21 \\
    &                                & SAGE        & 47.83 $\pm$ 5.38 & 46.56 $\pm$ 6.10 & \uline{44.01 $\pm$ 5.34} & \uline{41.42 $\pm$ 5.82} & \uline{37.67 $\pm$ 1.60} & 30.36 $\pm$ 4.15 & 29.21 $\pm$ 0.15 & 7.45 $\pm$ 0.03 \\
    &                                & GAT         & 54.10 $\pm$ 6.96 & 46.75 $\pm$ 8.28 & 38.50 $\pm$ 4.47 & 33.91 $\pm$ 5.20 & 37.24 $\pm$ 2.05 & \textbf{31.49 $\pm$ 3.81}& 25.95 $\pm$ 0.14 & 5.30 $\pm$ 0.02 \\ 
    \hline
    \multirow{5}{*}{\begin{tabular}[c]{@{}c@{}}Pre-training \&\\ Fine-tuning\end{tabular}} 
    & \multirow{2}{*}{Contrastive}   & GraphCL     & 43.20 $\pm$ 6.25 & 37.12 $\pm$ 8.78 & 36.73 $\pm$ 6.74 & 33.39 $\pm$ 7.21 & 33.88 $\pm$ 3.17 & 27.42 $\pm$ 4.51 & 12.51 $\pm$ 5.16 & 10.51 $\pm$ 2.36 \\
    &                                & DGI         & 41.16 $\pm$ 4.31 & 34.51 $\pm$ 4.15 & 38.22 $\pm$ 2.46 & 34.21 $\pm$ 4.42 & 34.02 $\pm$ 4.21 & 27.93 $\pm$ 3.25 & 12.87 $\pm$ 3.53 & 11.23 $\pm$ 5.32 \\
    \cdashline{3-11}
    & \multirow{3}{*}{\begin{tabular}[c]{@{}c@{}}Foundation\\Models\end{tabular}} 
     & GraphPrompt & 33.15 $\pm$ 6.52 & 16.24 $\pm$ 9.12 & 30.39 $\pm$ 6.24 & 11.65 $\pm$ 4.07 & 27.11 $\pm$ 1.35 & 14.22 $\pm$ 3.14 & 12.24 $\pm$ 7.76 & 11.42 $\pm$ 4.94 \\
    &                                & GCOPE       & 34.12 $\pm$ 1.24 & 27.05 $\pm$ 4.46 & 39.91 $\pm$ 6.74 & 35.90 $\pm$ 8.27 & 35.98 $\pm$ 1.66 & 25.43 $\pm$ 4.62 & 14.42 $\pm$ 3.21 & \uline{12.41 $\pm$ 5.25} \\
    &                                & SAMGPT      & 50.23 $\pm$ 9.57 & 46.12 $\pm$ 11.89& 38.48 $\pm$ 6.40 & 36.40 $\pm$ 6.94 & 34.56 $\pm$ 2.47 & 31.39 $\pm$ 3.33 & 14.68 $\pm$ 5.72 & 10.83 $\pm$ 2.87 \\ 
    \hline
    \multicolumn{2}{c|}{\textbf{Ours}} & \textbf{\model} & \textbf{66.41 $\pm$ 5.01} & \textbf{64.99 $\pm$ 5.94} & \textbf{85.18 $\pm$ 9.45} & \textbf{83.15 $\pm$ 11.43}& \textbf{37.84 $\pm$ 5.90} & \uline{31.45 $\pm$ 7.07} & 17.52 $\pm$ 4.14 & \textbf{15.03 $\pm$ 3.17} \\
    \hline\hline
    \end{tabular}}
\end{table*}

\begin{table*}[!htb]
  \centering
  \caption{Few-shot graph classification results. Average Micro-F1 and Macro-F1 (\%) $\pm$ standard deviation are reported. \textbf{Boldface} letters mark the best performance, while \uline{underlined} letters indicate the second best.}
  \renewcommand\arraystretch{1.3}
  \label{tab:extended_results}
  \resizebox{\textwidth}{!}{
    \begin{tabular}{cc|c|cc|cc|cc|cc}
    \hline\hline
    \multirow{2}{*}{\textbf{Paradigm}} & \multirow{2}{*}{\textbf{Category}} & \multirow{2}{*}{\textbf{Method}} & \multicolumn{2}{c|}{ACM} & \multicolumn{2}{c|}{DBLP} & \multicolumn{2}{c|}{IMDB} & \multicolumn{2}{c}{Freebase} \\
    \cline{4-11}
    & & & Micro-F1 & Macro-F1 & Micro-F1 & Macro-F1 & Micro-F1 & Macro-F1 & Micro-F1 & Macro-F1 \\
    \hline
    \multirow{5}{*}{\begin{tabular}[c]{@{}c@{}}Supervised\\ Learning\end{tabular}}
    & \multirow{2}{*}{Heterogeneous} & HAN         & 47.47 $\pm$ 8.42  & 43.02 $\pm$ 10.98 & \uline{49.13 $\pm$ 8.66}  & \uline{45.68 $\pm$ 9.38}  & 34.20 $\pm$ 2.99 & 26.48 $\pm$ 6.38 & \textbf{34.13 $\pm$ 7.39} & 7.67 $\pm$ 2.48 \\
    &                                & Simple-HGN  & 44.26 $\pm$ 5.16  & 40.02 $\pm$ 7.54  & 46.12 $\pm$ 6.26  & 43.15 $\pm$ 5.21  & 35.16 $\pm$ 3.15 & 27.34 $\pm$ 6.26 & \uline{31.73 $\pm$ 4.27} & 5.32 $\pm$ 1.29 \\
    \cdashline{3-11}
    & \multirow{3}{*}{Homogeneous}   & GCN         & \uline{54.68 $\pm$ 9.08}  & \uline{48.26 $\pm$ 11.79} & 43.38 $\pm$ 6.28  & 37.64 $\pm$ 7.06  & 35.02 $\pm$ 2.85 & 27.36 $\pm$ 4.89 & 20.04 $\pm$ 0.41 & 9.40 $\pm$ 0.05 \\
    &                                & SAGE        & 42.63 $\pm$ 5.59  & 36.44 $\pm$ 8.02  & 47.14 $\pm$ 8.32  & 43.64 $\pm$ 9.28  & \uline{35.24 $\pm$ 2.52} & \uline{28.80 $\pm$ 5.35} & 26.94 $\pm$ 4.41 & 9.87 $\pm$ 2.60 \\
    &                                & GAT         & 42.61 $\pm$ 4.17  & 34.01 $\pm$ 5.57  & 44.67 $\pm$ 7.53  & 40.61 $\pm$ 8.80  & 34.89 $\pm$ 2.63 & 26.89 $\pm$ 5.65 & 21.19 $\pm$ 5.75 & 8.78 $\pm$ 0.94 \\
    \hline
    \multirow{5}{*}{\begin{tabular}[c]{@{}c@{}}Pre-training \&\\ Fine-tuning\end{tabular}}
    & \multirow{2}{*}{Contrastive}   & GraphCL     & 43.87 $\pm$ 7.13  & 37.56 $\pm$ 9.99  & 39.98 $\pm$ 8.22  & 31.20 $\pm$ 8.69  & 35.04 $\pm$ 1.79 & 26.38 $\pm$ 5.45 & 12.31 $\pm$ 5.76 & 7.92 $\pm$ 2.35 \\
    &                                & DGI         & 40.21 $\pm$ 4.26  & 34.66 $\pm$ 6.48  & 36.41 $\pm$ 6.51  & 28.12 $\pm$ 4.21  & 34.15 $\pm$ 2.16 & 27.33 $\pm$ 6.15 & 11.24 $\pm$ 5.53 & 7.43 $\pm$ 1.89 \\
    \cdashline{3-11}
    & \multirow{3}{*}{\begin{tabular}[c]{@{}c@{}}Foundation\\Models\end{tabular}}
    & GraphPrompt & 31.16 $\pm$ 5.16  & 14.15 $\pm$ 6.16  & 28.56 $\pm$ 4.16  & 11.65 $\pm$ 4.07  & 24.16 $\pm$ 4.15 & 13.82 $\pm$ 3.25 & 10.83 $\pm$ 5.46  & 9.39 $\pm$ 4.75 \\
    &                                & GCOPE       & 33.15 $\pm$ 5.26  & 25.16 $\pm$ 3.16  & 35.43 $\pm$ 3.17  & 35.90 $\pm$ 8.27  & 34.17 $\pm$ 2.83 & 24.42 $\pm$ 3.62 & 11.42 $\pm$ 4.16 & \uline{10.11 $\pm$ 4.36} \\
    &                                & SAMGPT      & 42.88 $\pm$ 10.48 & 37.17 $\pm$ 10.69 & 40.28 $\pm$ 7.82  & 33.95 $\pm$ 9.27  & 34.58 $\pm$ 3.42 & 27.77 $\pm$ 5.41 & 13.58 $\pm$ 4.51 & 9.74 $\pm$ 2.31 \\
    \hline
    \multicolumn{2}{c|}{\textbf{Ours}} & \textbf{\model} & \textbf{57.86 $\pm$ 11.26} & \textbf{52.16 $\pm$ 14.32} & \textbf{55.72 $\pm$ 14.69} & \textbf{49.19 $\pm$ 15.06} & \textbf{36.01 $\pm$ 4.81} & \textbf{29.13 $\pm$ 6.23} & 17.16 $\pm$ 9.43 & \textbf{11.47 $\pm$ 3.36} \\
    \hline\hline
    \end{tabular}}
\end{table*}

\subsubsection{Evaluation Protocol and Few-Shot Settings}
To evaluate cross-domain transferability, a \textbf{Leave-One-Out} pre-training and fine-tuning protocol is adopted \cite{GCOPE, SAMGPT}. Specifically, one dataset is designated exclusively as the target domain, while the remaining datasets serve as source domains for multi-domain pre-training. For downstream few-shot adaptation, $K$ labeled nodes per category ($|C|$) are randomly sampled to form the support set $\mathcal{S}$ ($|\mathcal{S}| = |C| \times K$), with the remaining nodes evenly bisected into validation ($\mathcal{V}$) and test ($\mathcal{T}$) sets. During this phase, the pre-trained backbone is entirely frozen, and only the lightweight prompts are updated.

All baselines are evaluated using the same few-shot data partitioning strategy. For the homogeneous and heterogeneous GNNs, training is performed in a fully supervised manner on the same few-shot training set used during the fine-tuning stage of pre-training, contrastive, and graph foundation model baselines. For the remaining baselines, we follow the standard pre-training and fine-tuning paradigm: these methods are first pre-trained on the source domains under the leave-one-out protocol and then adapted to the target domain using the same few-shot training set. Furthermore, standard homogeneous GNNs, contrastive learning methods, and homogeneous foundation models are inherently designed for homogeneous structures. For these baselines, the input heterogeneous graphs are directly flattened into homogeneous topologies by ignoring node and edge type distinctions prior to execution.

\subsection{Analysis}
We answer five \textbf{R}esearch \textbf{Q}uestions (\textbf{RQ}) and demonstrate our arguments by extended experiments.

\subsubsection{\textbf{RQ1: Does \model surpass state-of-the-art baselines in text-free, few-shot cross-domain heterogeneous graph learning?}}

We evaluate \model against diverse baseline paradigms under the extreme 1-shot cross-domain setting. The node classification results are reported in Table~\ref{tab:main_results}, while the graph classification results are reported in Table~\ref{tab:extended_results}. Overall, the results consistently demonstrate the strong positive transferability of \model across both node-level and graph-level tasks. It is worth noting that GCN, GraphSAGE, GAT, HAN, and Simple-HGN are supervised baselines trained directly on the target few-shot set, whereas \model, GraphCL, DGI, GraphPrompt, GCOPE, and SAMGPT all follow the same pre-training-and-fine-tuning protocol under the leave-one-out setting. This distinction is important for interpreting the results: supervised baselines mainly reflect direct adaptation ability under scarce labels, while the latter group evaluates whether transferable cross-domain knowledge can be effectively acquired before target adaptation.

For node classification, \model achieves the best results on three of the four target datasets, with the clearest gains on ACM and DBLP. On ACM, it improves over the strongest baseline, GCN, by 8.8\% in Micro-F1 and 15.7\% in Macro-F1. On DBLP, the advantage becomes particularly pronounced, where \model substantially outperforms both the strongest supervised baseline, GraphSAGE, and the strongest foundation baseline, GCOPE, by a wide margin in both Micro-F1 and Macro-F1. On IMDB, although the margin becomes smaller, \model still achieves the best Micro-F1 and the second-best Macro-F1, indicating stable transferability even when competing methods perform similarly. Overall, these results suggest that semantic-preserving transformation and Tri-Prompt adaptation enable more effective transfer of heterogeneous structural priors than direct supervision, contrastive pre-training alone, or homogeneous flattening.

A similar pattern is observed for graph classification, confirming that the advantage of \model extends beyond node-level prediction. As shown in Table~\ref{tab:extended_results}, \model ranks first on ACM, DBLP, and IMDB in both Micro-F1 and Macro-F1. On ACM, it improves over GCN by 5.8\% in Micro-F1 and 8.1\% in Macro-F1. On DBLP, it surpasses HAN by 13.4\% and 7.7\%, respectively. On IMDB, the gains remain consistent but more modest, reaching 2.2\% in Micro-F1 and 1.2\% in Macro-F1 over the strongest competitor, GraphSAGE. These results indicate that \model can transfer not only node-level semantics but also graph-level structural patterns.

Freebase reveals an important boundary condition. Unlike ACM, DBLP, and IMDB, this dataset is inherently feature-poor, so after one-hot initialization and SVD alignment, the target graph is reduced to a largely structural identifier space with very weak feature semantics. Under such a setting, supervised baselines can more easily overfit to local structural cues in the target few-shot set, whereas pre-training-and-fine-tuning methods are required to transfer knowledge across a much larger semantic gap. This helps explain why \model does not outperform HAN on Freebase Micro-F1, even though it still achieves the best Macro-F1 on both node classification and graph classification. Therefore, the Freebase results should be interpreted not simply as weaker overall transfer, but as evidence that class-balanced transfer under severe feature degeneration remains a challenging case for all pre-training-based paradigms. Taken together, these findings show that \model delivers consistent and often substantial improvements across both node classification and graph classification, validating the effectiveness of preserving heterogeneous semantics throughout transformation, pre-training, and fine-tuning.

\subsubsection{\textbf{RQ2: How resilient is \model to varying degrees of data scarcity during cross-domain adaptation?}}

\begin{figure}[!htb]
  \centering
  \includegraphics[width =\linewidth]{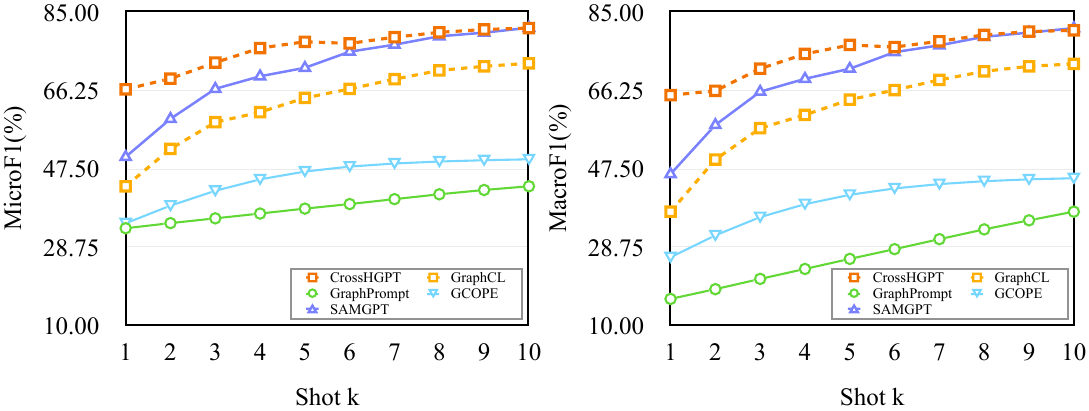}
  \caption{Impact of sample size ($k$) on ACM classification performance}
  \label{fig:kshot}
\end{figure}

To examine the robustness of \model under different levels of target-domain supervision, we conduct a $K$-shot analysis on ACM with $K \in \{1,2,\dots,10\}$. We compare \model with representative baselines under the ``pre-training and fine-tuning'' paradigm, and report the Micro-F1 and Macro-F1 trends in Fig.~\ref{fig:kshot}. Overall, \model consistently achieves the best performance across the entire $K$-shot range. Its advantage is particularly evident in the extremely low-shot regime ($K \leq 3$), where only a handful of labeled nodes are available for adaptation. This suggests that \model can exploit scarce supervision more effectively than competing methods, benefiting from transferable heterogeneous priors learned during pre-training rather than relying heavily on target-domain labels.

\begin{figure}[!htb]
  \centering
  \includegraphics[width=\linewidth]{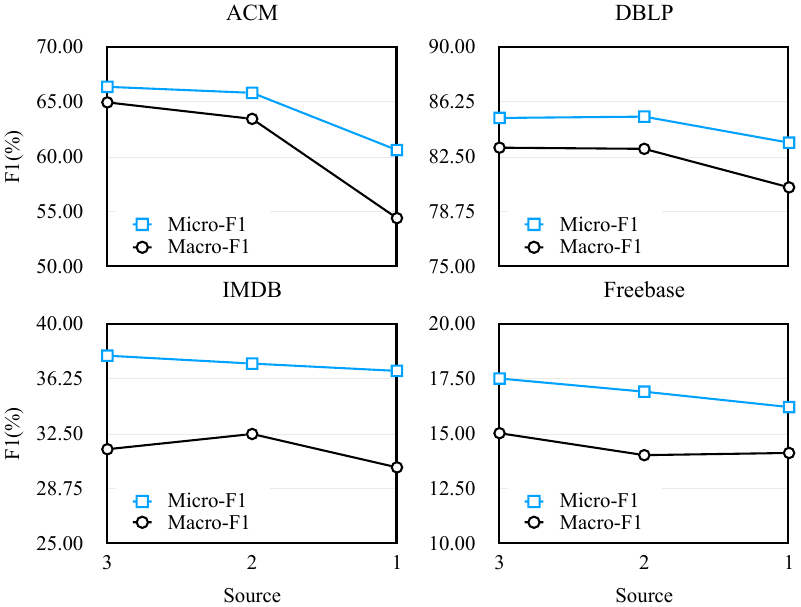}
  \caption{Performance comparison under different pre-training source settings. The full multi-source setting consistently achieves the best overall performance, while reducing the number of source domains leads to clear performance degradation on several target datasets.}
  \label{fig:source_ablation}
\end{figure}

As $K$ increases, all methods improve with more labeled data, but \model maintains a clear lead and exhibits a steady upward trend, reaching 81.06\% Micro-F1 and 80.52\% Macro-F1 at 10-shot. These results indicate that \model not only excels under extreme label scarcity, but also retains strong adaptation capability as supervision becomes less limited. In other words, preserving heterogeneous semantics during pre-training and fine-tuning yields stronger low-shot resilience and superior sample efficiency in cross-domain transfer.

\subsubsection{\textbf{RQ3: How important is multi-source pre-training for \model?}}

To investigate the effect of pre-training source domains, we progressively remove source datasets during pre-training while keeping the target-domain fine-tuning and evaluation protocol unchanged. Fig.~\ref{fig:source_ablation} shows that the full multi-source setting consistently yields the strongest overall performance, confirming that \model benefits from diverse source-domain knowledge during pre-training. When only one source domain is removed, the performance drop is relatively limited, especially on DBLP, indicating that the contribution of different source domains is not uniform across target domains. This suggests that different source domains provide transferable structural and semantic priors with different degrees of relevance to each target dataset.

When the pre-training setting is further reduced to a single source domain, the degradation becomes much more evident. In particular, the average Micro-F1 drop reaches 8.03\% on ACM and 1.98\% on DBLP. Overall, these results show that although individual source domains contribute unevenly, multi-source pre-training exposes \model to more diverse heterogeneous structures and relation patterns, enabling it to learn more transferable semantic priors and achieve better generalization on unseen target domains.

\begin{table*}[!htb]
  \centering
  \caption{Ablation study results of Tri-Prompt components on four target datasets. Boldface denotes the best performance.}
  \renewcommand\arraystretch{1.3}
  \label{tab:ablation}
  \resizebox{\textwidth}{!}{
    \begin{tabular}{c|cc|cc|cc|cc} 
    \hline\hline
    \multirow{2}{*}{\diagbox[width=11em, height=2.8em]{\textbf{Method}}{\textbf{Dataset}}} & \multicolumn{2}{c|}{ACM} & \multicolumn{2}{c|}{DBLP} & \multicolumn{2}{c|}{IMDB} & \multicolumn{2}{c}{Freebase} \\ 
    \cline{2-9}
    & Micro-F1 & Macro-F1 & Micro-F1 & Macro-F1 & Micro-F1 & Macro-F1 & Micro-F1 & Macro-F1 \\ 
    \hline
    w/o Edge Prompt       & 64.27 $\pm$ 8.60 & 59.40 $\pm$ 10.62 & 85.03 $\pm$ 9.60 & 83.06 $\pm$ 11.78 & 36.65 $\pm$ 6.14 & 31.37 $\pm$ 7.26 & 15.23 $\pm$ 6.08 & 9.69 $\pm$ 2.47 \\
    w/o Structure Prompt  & 62.72 $\pm$ 6.25 & 56.46 $\pm$ 8.01  & 84.87 $\pm$ 8.42 & 84.61 $\pm$ 10.16 & 36.37 $\pm$ 5.34 & 31.30 $\pm$ 6.41 & 15.02 $\pm$ 6.72 & 9.50 $\pm$ 2.49 \\
    w/o Feature Prompt    & 60.85 $\pm$ 6.60 & 54.56 $\pm$ 8.22  & 84.92 $\pm$ 8.57 & 84.03 $\pm$ 10.74 & 36.75 $\pm$ 5.98 & 31.27 $\pm$ 7.18 & 15.11 $\pm$ 5.92 & 9.93 $\pm$ 2.46 \\
    \hline
    \textbf{\model} & \textbf{66.41 $\pm$ 5.01} & \textbf{64.99 $\pm$ 5.94} & \textbf{85.18 $\pm$ 9.45} & \textbf{83.15 $\pm$ 11.43} & \textbf{37.84 $\pm$ 5.90} & \textbf{31.45 $\pm$ 7.07} & \textbf{17.52 $\pm$ 4.14} & \textbf{15.03 $\pm$ 3.17} \\
    \hline\hline
    \end{tabular}}
\end{table*}

\begin{figure*}[!ht]
    \centering
    \includegraphics[width = \linewidth]{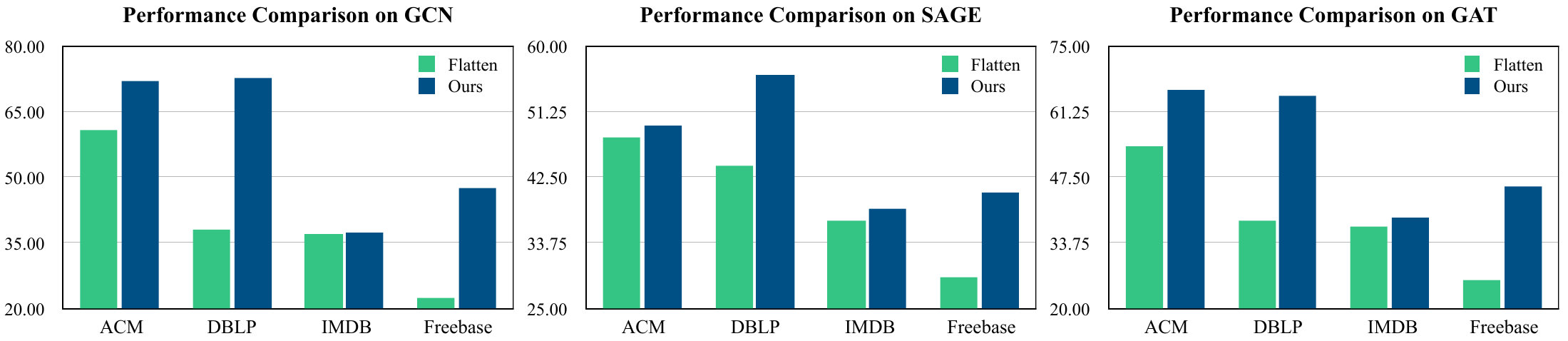}
    \caption{Comparison between direct flattening and our semantic-preserving graph transformation on GCN, GraphSAGE, and GAT under the 1-shot supervised setting, where the remaining nodes are split into validation and test sets following the protocol used in the main results.}
    \label{fig:transformation_bar}
\end{figure*}

\subsubsection{\textbf{RQ4. Does the Tri-Prompt in \model work?}}

Table~\ref{tab:ablation} reports the ablation results of the three prompt components. Overall, the full \model achieves the best performance on all four target datasets, showing that the gains of \model arise from the complementary effects of the entire Tri-Prompt design rather than any single prompt alone. Among the three variants, removing the Feature Prompt causes the largest degradation on ACM, where Micro-F1 decreases by 8.37\%. This indicates that feature-level calibration is particularly important when the target graph contains informative node attributes. Meanwhile, removing the Edge Prompt or the Structure Prompt also consistently weakens performance, especially on DBLP and IMDB, indicating that both relation-aware interactions and topology-aware adaptation are important for preserving transferable heterogeneous semantics.

We also observe that different datasets show different sensitivities to prompt removal. ACM is more affected by removing the Feature Prompt, whereas DBLP and IMDB exhibit relatively consistent declines across all variants. These results suggest that the three prompts capture complementary aspects of transferable knowledge, and their joint use is necessary for robust cross-domain adaptation.

\subsubsection{\textbf{RQ5: How crucial is the Semantic-Preserving Graph Transformation for retaining multi-relational priors?}}

To validate the contribution of the graph transformation module, we compare our semantic-preserving strategy with direct flattening using GCN, GraphSAGE, and GAT under the same 1-shot supervised setting. Specifically, for each target dataset, we use 1-shot labeled nodes for training and split the remaining nodes into validation and test sets following the same protocol as in the main results. As shown in Fig.~\ref{fig:transformation_bar}, the models equipped with our transformation consistently outperform their directly flattened counterparts across all three backbones.

This result indicates that directly flattening a heterogeneous graph discards important node- and edge-type semantics, making it harder for the model to exploit transferable structural priors under scarce supervision. In contrast, our transformation preserves multi-relational information in a unified homogeneous graph, allowing even standard homogeneous GNNs to better capture heterogeneous dependencies. These results suggest that semantic-preserving transformation provides an effective structural inductive bias for few-shot learning on heterogeneous graphs.

\section{Conclusions and Outlook}
In this work, we investigate the problem of text-free, few-shot, cross-domain learning on heterogeneous graphs and propose \model, a foundation framework that explicitly preserves and transfers heterogeneous structural semantics. By integrating semantic-preserving graph transformation, prompt-aware multi-domain pre-training, and parameter-efficient cross-domain fine-tuning, \model effectively mitigates heterogeneity shifts across domains without relying on external textual supervision.
Extensive experiments on four real-world heterogeneous graph benchmarks demonstrate that \model consistently achieves competitive and often superior performance under extremely limited target-domain supervision. The empirical results further indicate that preserving multi-relational semantics during both pre-training and adaptation is important for effective knowledge transfer in heterogeneous settings. In particular, the results of source-domain ablation, prompt ablation, and graph transformation analysis collectively support the effectiveness of the proposed design.
This work suggests that semantic-preserving transformation and prompt-based adaptation offer a practical direction for heterogeneous graph foundation models in text-free scenarios. In future work, it would be valuable to extend this framework to graph-level tasks, larger-scale graphs, and more challenging open-world scenarios, as well as to explore more efficient pre-training objectives and more general semantic alignment strategies across highly diverse domains.

\bibliographystyle{IEEEtran}
\bibliography{Reference,IEEEabrv}

\vspace{-20pt}
\begin{IEEEbiography}[{\includegraphics[width=1in,height=1.25in,clip,keepaspectratio]{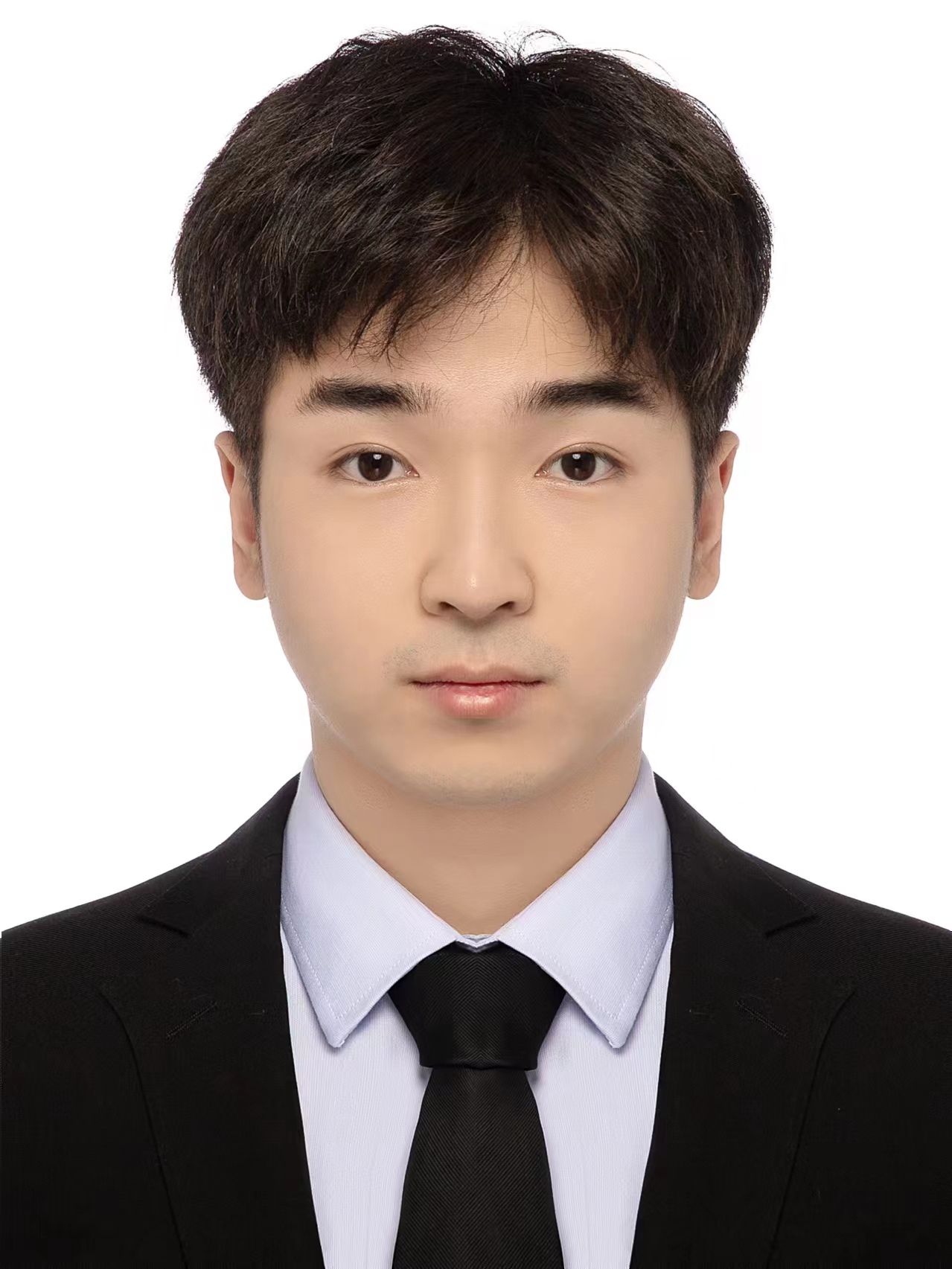}}]{Xuanze Chen}
	received the B.S. degree from Wenzhou University, Wenzhou, China, in 2023. He is currently pursuing the Ph.D. degree with the Institute of Cyberspace Security, Zhejiang University of Technology, Hangzhou, China. His current research interests include graph data mining and graph foundation models.
\end{IEEEbiography}

\vspace{-20pt}
\begin{IEEEbiography}[{\includegraphics[width=1in,height=1.25in,clip,keepaspectratio]{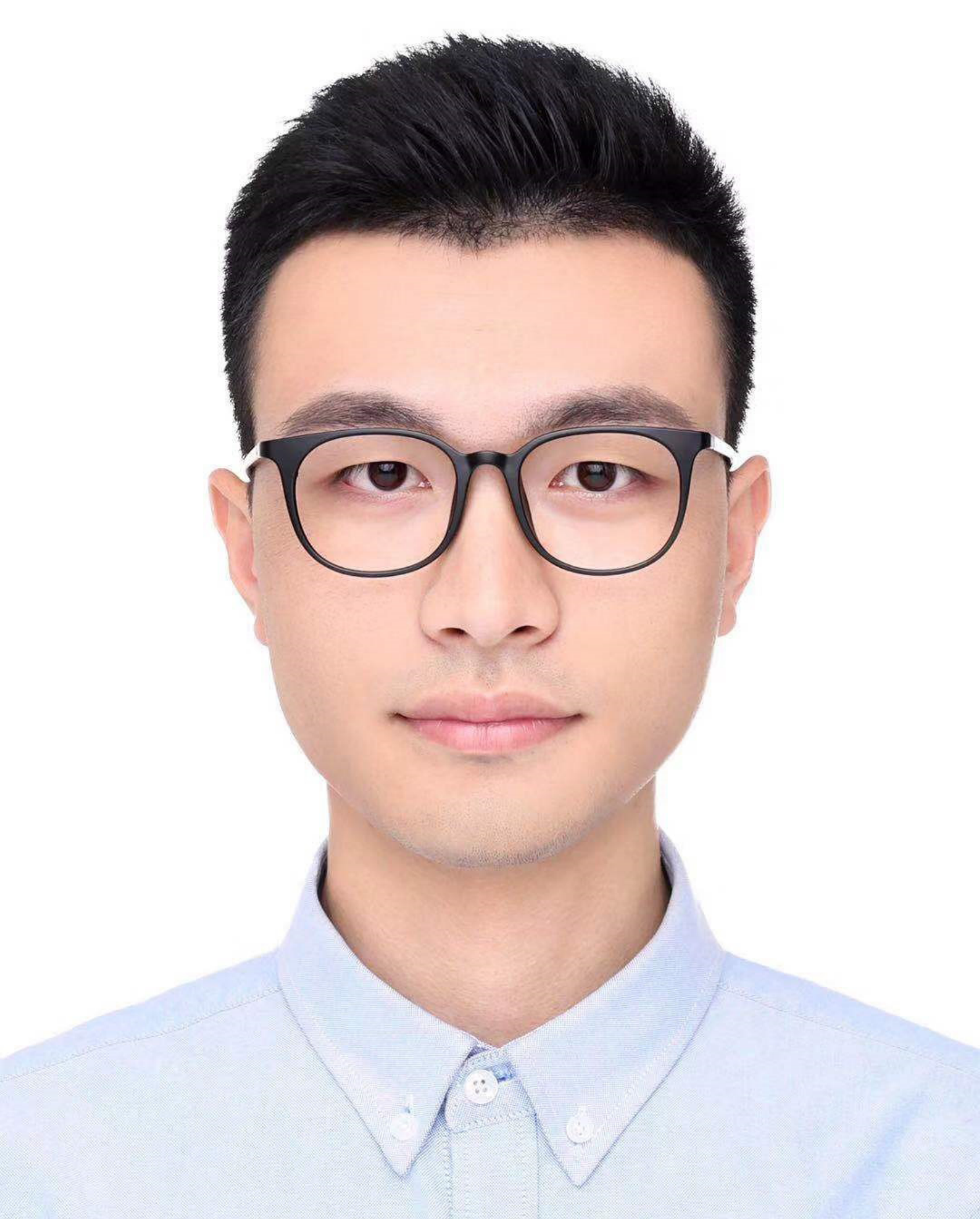}}]{Jiajun Zhou}
	received the Ph.D degree in control theory and engineering from Zhejiang University of Technology, Hangzhou, China, in 2023. He is currently a Postdoctoral Research Fellow with the Institute of Cyberspace Security, Zhejiang University of Technology. His current research interests include graph data mining, cyberspace security and data management.
\end{IEEEbiography}
\vspace{-20pt}

\begin{IEEEbiography}[{\includegraphics[width=1in,height=1.25in,clip,keepaspectratio]{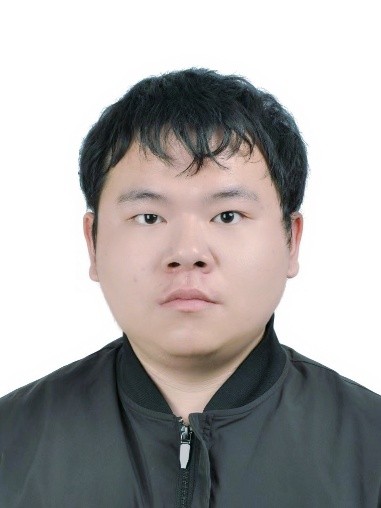}}]{Yadong Li}
	received a BS degree from Tianjin University of Technology, Tianjin, China, in 2024. He is currently pursuing a Master's degree in Control Science and Engineering at Zhejiang University of Technology. His current research interests include graph neural networks and large language models.
\end{IEEEbiography}
\vspace{-20pt}

\begin{IEEEbiography}[{\includegraphics[width=1in,height=1.25in,clip,keepaspectratio]{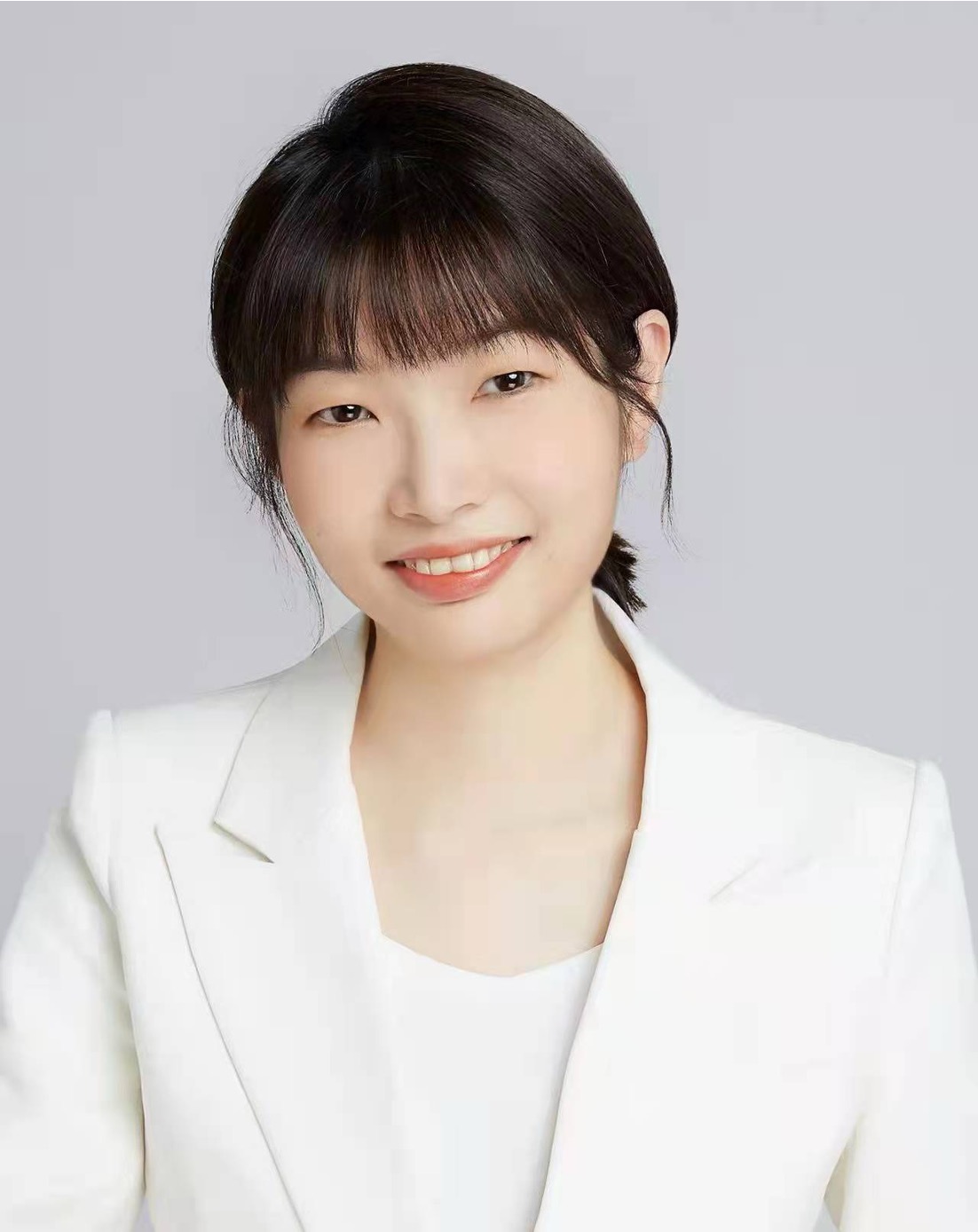}}]{Shanqing Yu}
	received the M.S. degree from the School of Computer Engineering and Science, Shanghai University, China, in 2008 and received the M.S. degree from the Graduate School of Information, Production and Systems, Waseda University, Japan, in 2008, and the Ph.D. degree, in 2011, respectively. She is currently a Lecturer at the Institute of Cyberspace Security and the College of Information Engineering, Zhejiang University of Technology, Hangzhou, China. Her research interests cover intelligent computation and data mining.
\end{IEEEbiography}
\vspace{-20pt}

\begin{IEEEbiography}[{\includegraphics[width=1in,height=1.25in,clip,keepaspectratio]{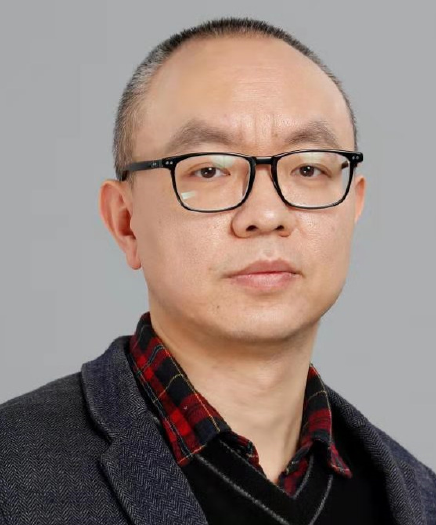}}]{Qi Xuan}(M'18) 
  received the BS and PhD degrees in control theory and engineering from Zhejiang University, Hangzhou, China, in 2003 and 2008, respectively. He was a Post-Doctoral Researcher with the Department of Information Science and Electronic Engineering, Zhejiang University, from 2008 to 2010, respectively, and a Research Assistant with the Department of Electronic Engineering, City University of Hong Kong, Hong Kong, in 2010 and 2017. From 2012 to 2014, he was a Post-Doctoral Fellow with the Department of Computer Science, University of California at Davis, CA, USA. He is a senior member of the IEEE and is currently a Professor with the Institute of Cyberspace Security, College of Information Engineering, Zhejiang University of Technology, Hangzhou, China. His current research interests include network science, graph data mining, cyberspace security, machine learning, and computer vision.
\end{IEEEbiography}

\end{document}